\pdfoutput=1

\documentclass[11pt]{article}

\usepackage{acl}

\usepackage{times}
\usepackage{latexsym}

\usepackage[T1]{fontenc}

\usepackage[utf8]{inputenc}

\usepackage{microtype}

\usepackage{inconsolata}

\usepackage{graphicx}
\usepackage{booktabs}
\usepackage[most]{tcolorbox}

\title{The InviTE Corpus: Annotating Invectives in Tudor English Texts for Computational Modeling}

\author{Sophie Spliethoff$^{1}$, Sanne Hoeken$^{2}$, Silke Schwandt$^1$, Sina Zarrieß$^2$ \and Özge Alaçam$^2$\\$^1$Dept. of History, Faculty of History, Philosophy and Theology, Bielefeld University\\$^2$Computational Linguistics, Dept. of Linguistics, Bielefeld University\\
\texttt{\{sophie\_jasmin.spliethoff, sanne.hoeken, silke.schwandt,}\\\texttt{sina.zarriess, oezge.alacam\}@uni-bielefeld.de} \\}

\begin{document}
\maketitle
\begin{abstract}
In this paper, we aim at the application of Natural Language Processing (NLP) techniques to historical research endeavors, particularly addressing the study of religious invectives in the context of the Protestant Reformation in Tudor England. We outline a workflow spanning from raw data, through pre-processing and data selection, to an iterative annotation process. As a result, we introduce the InviTE corpus -- a corpus of almost 2000 Early Modern English (EModE) sentences, which are enriched with expert annotations regarding invective language throughout 16\textsuperscript{th}-century England. Subsequently, we assess and compare the performance of fine-tuned BERT-based models and zero-shot prompted instruction-tuned large language models (LLMs), which highlights the superiority of models pre-trained on historical data and fine-tuned to invective detection. 

\end{abstract}

\section{Introduction}

Bridging theoretical frameworks on historical language with computational approaches requires operationalization and adaptation. Contemporary Natural Language Processing (NLP) typically relies on large-scale datasets annotated with relatively simple schemes (most often binary or few-class categories), which facilitates scaling, consistency, and computational efficiency. Traditional historical research usually involves a close-reading analysis of selected sources in order to trace, understand, and contextualize historical developments. An interdisciplinary approach that includes the application of NLP models requires that historical analysis be translated into systematic categories. Although challenging, utilizing NLP techniques offers an innovative opportunity to efficiently integrate a wide range of historical sources simultaneously, which is of increasing interest, particularly with regards to the recent efforts to digitize large historical corpora, which can no longer be surveyed in their full entirety through qualitative analysis alone.

\begin{figure}[h!]
    \centering
    \includegraphics[width=0.8\columnwidth]{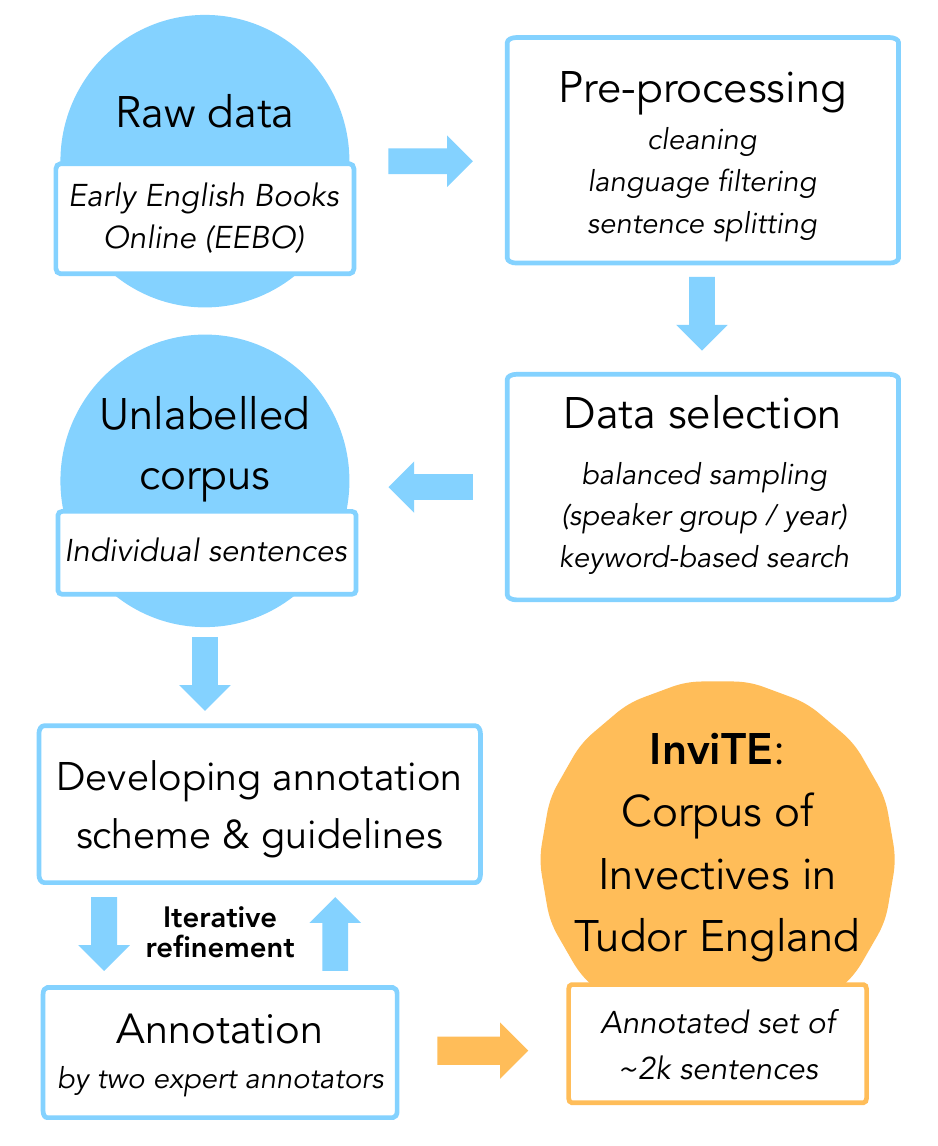}
    \caption{Illustration of the InviTE corpus creation process, highlighting the main steps taken from raw data to a structured and annotated dataset.}
    \label{fig:placeholder}
\end{figure}

In this study we focus on invectives in Early Modern English (EModE) texts published throughout the Tudor era (1485-1603) in England. At this time, medial, societal and religious upheavals paved the way for a vivid public discourse, which revolved primarily around theological controversies. 

In contrast to modern notions of hate speech, for example determined by the incitement of "violence or hate against groups, based on specific characteristics" by \citet{fortuna_survey_2018}, the framework of invectivity proposed by \citet{schwerhoff_invektivitat_2017} entails a broader definition, stating that all forms of communications, which potentially derogate an opponent group or individual are invective. Moreover, it is crucial to take into account that invectives in pre-modern sources are situated in a markedly different historical and cultural context. 

Operationalizing historical research frameworks for computational purposes requires careful adaptation: enriching annotation schemes and building detailed annotation guidelines, incorporating metadata such as author and publication year, and developing methods to deal with sparse and historically variable data.

In this paper, we present the following contributions:
\begin{enumerate}
    \item We introduce a corpus-building and annotation methodology designed to operationalize historical theories of invective language for computational purposes (Figure \ref{fig:placeholder}).
    \item Based on this, we present the InviTE corpus, a sentence-level corpus of Invectives in Tudor England, richly annotated by domain experts for invectivity using a fine-grained scheme tailored to EModE texts.
    \item We evaluate and compare the performance of fine-tuned BERT models and recent instruction-tuned Large Language Models (LLMs) on the task of invective detection and show that fine-tuning is still crucial: even small models, once fine-tuned, outperform much larger LLMs used in zero-shot settings.
\end{enumerate}

By combining expertise from history and NLP, this work demonstrates how conceptual frameworks from the humanities can be operationalized for computational research. The results of our computational experiments underline the value of corpus-building and annotation work as fine-tuning proved to be the most successful modeling approach.


\section{Related work}

In this section, we outline previous approaches to both investigating historical invectives in the humanities and applying computational (language) models to historical data.

\subsection{Studying Early Modern English invectives}
\label{historical-invectives}
Investigating invectivity in Early Modern England has long been a relevant research field for historians, theologians and literary scholars. This mostly includes studies on (religious) polemics and rhetorics, exploring historical strategies, motivations and effects of disparaging language and lines of argumentation across various historical epochs and cultural spheres. These works usually investigate polemical writings of selected authors (e.g. \citet{kelly_chasing_2015}) or genres (e.g \citet{bevan_zlatar_reformation_2011, suerbaum_language_2015}) and provide detailed analyses of linguistic means and rhetorical structures in their specific historical, social and cultural contexts. 


However, a systematic and comprehensive approach to all these terms and concepts has been missing until \citet{schwerhoff_invektivitat_2017} put forward a research framework, which serves as the base of our study. Within this framework, all (speech) acts which potentially disparage an individual or group as invectives, are defined as invective, which allows for a more comprehensive analysis.

\subsection{Historical research meets computational methods}

The operationalization of historical conceptual frameworks for computational study introduces several challenges. Digitized collections like Early English Books Online (EEBO) enable large-scale computational studies, but transcription errors, OCR noise, and highly variable spelling complicate text processing \citep{vanstrien-etal-2020, todorov-colavizza-2022}. Approaches such as automatic normalization mitigate some of these issues, though they remain partial solutions and risk obscuring historically meaningful orthographic variation \citep{bollmann-2019-large}.

Beyond pre-processing, historical text analysis focused on concepts like invective language requires annotation strategies that are adapted to these interpretation-heavy phenomena. Even for contemporary text, annotating evaluative language often proves challenging due to subjectivity and context dependence \citep{plank-2022-problem}. This difficulty is further amplified in historical settings. Previous work on sentiment analysis in historical corpora, for instance, reports moderate annotator agreement due to not only the effect of subjectivity, but also of cultural value change over time \citep{sprugnoli2016towards, allaith-etal-2023-sentiment}. 

A directly related contribution is the work by \citet{hoeken_towards_2023}, who introduce a dataset of 16\textsuperscript{th}-century religious texts annotated for lexical semantic shifts and hateful content (on the word-level). Their findings likewise highlight the complexity of annotating pre-modern polemical language. Building on these insights, our study incorporates an annotation scheme that explicitly records annotator confidence to represent interpretative uncertainty. Additionally, whereas \citet{hoeken_towards_2023} evaluate their dataset solely with the MacBERTh model, we extend the comparison to include multiple BERT-based models as well as recent LLMs. The following subsection discusses these models in more detail.


\subsection{NLP approaches to historical language}

In recent years, specialized language models for historical text have emerged. Domain adaptation through retraining on historical corpora has proven effective. MacBERTh \citep{MacBERTh_2021} was one of the first BERT variants trained on English texts from 1450 to 1900 and outperformed contemporary models on historical tasks \citep{manjavacas_adapting_2022}. Related efforts include HistBERT for 20\textsuperscript{th}-century English \citep{qiu2022histbert}, GysBERT for historical Dutch \citep{manjavacas-arevalo-fonteyn-2022-non}, and GHistBERT for historical German \citep{beck-kollner-2023-ghisbert-training}. These models have been evaluated on relatively well-defined tasks such as POS tagging and Word Sense Disambiguation. 
Such tasks do not capture the interpretative complexity (as discussed above) or sparsity characteristic of invective language.

More recently, LLMs have begun to be applied to historical contexts. In contrast to specialized models that are retrained or fine-tuned on historical corpora, these models, which contain billions of parameters, are trained on massive data collections to develop broad language and reasoning abilities. A common way to use these models is through prompting, where tasks are specified with instructions or examples rather than modifying the model itself. 
This approach offers flexibility across diverse tasks, but existing studies have not yet demonstrated consistently convincing results for historical language phenomena. For example, \citet{zhang-etal-2024-crosstemporal} explore cross-temporal summarization with GPT-3.5, and \citet{cassotti-tahmasebi-2025-sense} evaluate generative models on producing historically accurate word usages. Other studies assess LLM performance in a historical knowledge task \citep{hauser2024large} or pragmatic annotation of Early Modern English \citep{huang2025assessing}. 
These studies highlight that LLMs are not yet reliable stand-alone tools for historical research, but they can take a complementary role in tasks that benefit from iterative refinement and scholarly oversight.


\section{Data}

The following outlines the methodological steps, informed by historical research, for building the InviTE corpus (cf. Figure \ref{fig:placeholder}).

\subsection{Historical context}
We sourced the data from \textit{Early English Books Online} (EEBO), selecting all available EModE printed works which were published in England between 1485 and 1603. This time period captures the Protestant Reformation in England and reflects the public discourse, mainly revolving around religious questions and controversies. 

With the introduction of the printing press in England in 1476, the production of books was revolutionized. Authors, printers and publishers were able to produce books and pamphlets faster, cheaper, and in larger numbers, while reading audiences grew in numbers because books and pamphlets became more affordable. This allowed authors to respond to contemporary social, political, and religious events (for an introduction to printed sources in Early Modern England, see e.g. \citet{raven_renaissance_2023} and \citet{green_print_2000}). As highlighted by \citet{kastner_jedermann_2020}, the Reformation era was particularly characterized by religious controversies carried out in public, we we can expect that the selected sources contain a significant amount of invectives.


The texts included in this data set are highly diverse as they came in various formats and genres, uniting different styles and conventions of writing. Additionally, the selected period covers the transition from Middle English (ME) to EModE. Therefore, the English language was in constant change and had not yet been standardized (cf. \citet{gramley_history_2024}).

\subsection{Pre-processing and data selection}
\label{subsec:cleaning-corpus}
To clean the files, we removed headers and footers, blank lines, and page numbers. The sentences were split and assigned with unique sentence IDs. As EModE texts often include passages written in other languages, such as Latin, we used langdetect to only keep sentences containing EModE. 

From the cleaned source texts, we compiled a corpus of 1975 sentences. 1443 sentences are randomly selected with an equal distribution according to publication year and speaker group (catholic and non-catholic authors). Information on the authors' genders and religious confessions were manually added, while all other metadata were provided by EEBO. The initial annotation phase revealed that the corpus did not contain an adequate number of invective sentences (182 out of 1443) to support effective training of classification models. In order to mitigate this imbalance in the data, 532 sentences were manually added through keyword-based search. We based the selection of keywords on the top hateful terms detected in a smaller 16\textsuperscript{th}-century corpus (see \citet{hoeken_towards_2023}) and on terms we considered potentially invective among the most frequent terms across the whole corpus -- for a list of the keywords used, see Appendix \ref{sec:app-keywords}.

\section{Annotation}

\begin{table*}[th!]
\centering
\resizebox{0.7\textwidth}{!}{%
\begin{tabular}{lclclclc}
\toprule
\textbf{Invectivity} & \textbf{n} & \textbf{Type} & \textbf{n} & \textbf{Target} & \textbf{n} & \textbf{Confidence} & \textbf{n}\\
\midrule
Non-invective & 1415 & Non-invective & 1415 & Non-invective & 1415 & Confident & 1860 \\
Invective & 560 & Literal & 330 & Sinful behavior & 181 & Less confident & 115 \\
&  & Metaphorical & 218 & Confession & 107 & &\\
&  & Uncertain & 12 & Religious belief & 100 & &\\
&  & & & Political-religious & & &\\
&  & &  & misconduct & 93 & &\\
&  & &  & Other & 79 & &\\
\bottomrule
\end{tabular}
}
\caption{Statistics of the final annotated dataset.}
\label{tab:ann-scheme-stats}
\end{table*}

Based on the definition of invectivity of \citet{schwerhoff_invektivitat_2020} presented in Section \ref{historical-invectives}, the data set has been annotated by two domain expert annotators trained in theory-driven interpretation of historical, cultural, and religious conflicts through previous historical and political studies. 

\subsection{Iterative labeling method}
\label{subsec:iterative}
In order to make sure that the sentences are coherently annotated, we set up guidelines including information on the historical sources, the historical context, annotation categories and guiding rules for difficult cases following \citet{reiter_anleitung_2020}, who provides a guide for annotations in digital humanities research. 

With an initial annotation scheme, we focused on categories of social difference: \textit{gender}, \textit{status}, \textit{origin}, \textit{religious confession}, and \textit{other}. However, after the first annotation phase, we found that a consistent annotation is too challenging, as invectives may be based on more than one of these categories at the same time. Taking these intersections into account would result in an overly complex annotation scheme. Secondly, we observed that due to the fact that a large part of the source texts were published in the context of the Protestant Reformation, most invectives were uttered for religious reasons, while the category \textit{religious confession} was too narrow to cover the range of religion-based invectives.
Thirdly, it became clear that in a random selection of sentences, we might not find enough invectives. To mitigate this problem, we included an additional set of sentences through a keyword-based approach as described in Section \ref{subsec:cleaning-corpus}.

\subsection{Final annotation scheme}
By the above-described iterative process, we obtain an annotation scheme that allows for both coherent, reliable labelling of our corpus, which is outlined in the following (for a tabular overview, see Table \ref{tab:ann-scheme-stats}). 
\paragraph{Invectivity.}  The data is labelled according to whether a sentence contains invective language according to the before-mentioned definition based on \citet{schwerhoff_invektivitat_2017}. All sentences, which do not contain invective language, are labelled \textit{non-invective} -- this label is directly utilized as sub-category for the subsequent target and type annotations.

\paragraph{Type.} If the sentence contains invectives, they are labelled according to whether the invective part is conveyed in a \textit{literal} or \textit{metaphorical} way. 
Metaphors are omnipresent and play a crucial role both in religious writing and the study thereof -- for example, the collaborative reseach center (CRC) 1475\footnote{https://sfb1475.ruhr-uni-bochum.de/} was recently formed to examine the subject of religious metaphor in detail. 

\paragraph{Target.} Introducing more differentiated sub-categories, the focus is shifted towards the distinction of religious invectives. They include:
\begin{itemize}
    \item \textit{sinful behavior}: An act against moral or divine law is inveighed against.
    \item \textit{political-religious misconduct}: An act against moral or divine law is inveighed against, which is directly associated with a specific political and/or religious conviction, (i.e. treason).
    \item \textit{religious belief}: A person or group is inveighed against based on false and/or erring faith or paganism.
    \item \textit{confession}: A person or group is inveighed against because of their confession. 
\end{itemize}
Invectives which do not target religious attributes or behaviour are labelled \textit{other}. 

\paragraph{Confidence.} The two sub-categories \textit{confident} and \textit{less confident} indicate the annotator's confidence only with regards to the decision whether the sentence contains invective language or not.

\subsection{Annotation process} 

Subsequently, the selected corpus of 1975 sentences was initially annotated by one annotator, with 1675 of these sentences also labeled by a second annotator (an example can be viewed in Appendix \ref{sec:app-ann}). In addressing the subjectivity inherent to invective language, we followed a rather prescriptive paradigm \citep{rottger_two_2022} and established a shared definition of invectives and integrated guidance on the Reformation context, the type of data, and the historical dimension of the texts within the annotation guidelines. Nevertheless, subjective bias inherent to human annotation cannot be entirely avoided, and interpreting historical sources adds an additional layer of uncertainty. As \citet{craig_reformation_2016} notes, it is difficult even for trained historians to interpret pre-modern instances of irony, humor and polemics correctly. 

\subsection{The InviTE Corpus}

Our final data set, the InviTE corpus, contains 1975 sentences derived from a variety of different EModE texts, many of them related to the Reformation discourse in Tudor England. The sentences are enriched with (I) additional metadata regarding text title, publication year and author along with their genders and religious confessions and with (II) annotations with regards to invective language (1675 of the sentences are double-annotated). The corpus now includes 1415 sentences labelled as invective and 560 sentences labelled as non-invective. The frequency statistics for all remaining annotated categories are provided in Table \ref{tab:ann-scheme-stats}.


\paragraph{Inter-annotator agreement.} We assess inter-annotator agreement (IAA) by comparing the labels provided by the two annotators across all three annotation categories (\textit{invective}, \textit{type}, and \textit{target}). For each category, we report both raw percent agreement and Cohen’s Kappa ($\kappa$); the latter corrects for chance agreement. The scores indicate moderate agreement for the binary \textit{invective} label (87\%, $\kappa$ = 0.53) and \textit{type} (85\%, $\kappa$ = 0.48), and fair agreement for \textit{target} (82\%, $\kappa$ = 0.39). 
This pattern of decreasing agreement across categories is expected as both \textit{type} and \textit{target} are conditionally dependent on the identification of invective: if annotators disagree on whether an instance is invective, disagreement on its type or target necessarily follows. So, agreement on \textit{invective} forms an upper bound for the other categories. Additionally, while \textit{invective} is a binary decision, \textit{type} and \textit{target} are multi-class variables, which increases the likelihood of disagreement between annotators. 

To further explore annotation agreement, we examined the relationship between annotators' self-reported confidence (confident vs. less confident, provided only for the invective label) and agreement outcomes. Agreement was reached in 1456 cases, where the average confidence across both annotators was very high (97\%), while disagreement occurred in 219 cases, where average confidence dropped to 87\%. This pattern indicates that disagreements are associated with lower annotator certainty and validates the usefulness of the confidence measure.


\section{Computational methods}

Having operationalized the historical framework of invective language and developed the corpus, we can now evaluate how well computational models capture these phenomena, which represents the ultimate goal of our interdisciplinary approach.
We conduct experiments to evaluate automatic invective detection in EModE texts. The classification task is defined as follows: given a sentence from the corpus, predict whether it contains invective language or not. 

\subsection{Fine-tuned BERT-models}

We adopt a common approach to supervised text classification and fine-tune pretrained transformer-based language models on the task of automatic invective detection in EModE texts. Specifically, we experimented with three models: \texttt{BERT-base}, the original English BERT model trained on general-domain corpora \citep{devlin-etal-2019-bert}; \texttt{XLM-RoBERTa-large}, a large multilingual model trained on 100 languages \citep{conneau-etal-xlmr}; and \texttt{MacBERTh}, a BERT-based model pretrained on historical English \citep{MacBERTh_2021}. Including \texttt{BERT-base} and \texttt{XLM-RoBERTa-large} allows us to evaluate the effects of pretraining data, domain adaptation and model scale on invective detection performance. These models are encoder-based transformers, meaning that they generate contextualized representations of input text, in contrast to decoder-only architectures typically used in LLMs, which generate text autoregressively.

Each model was trained and evaluated using 10-fold cross-validation, where each fold used 80\% of the data for training, 10\% for validation, and 10\% for testing. To address class imbalance, the folds were generated using stratified sampling to maintain proportional representation of all classes in the training, validation, and test sets. During training, we employed a custom stratified batch sampler that preserves class balance within each mini-batch.
Additional implementation details are provided in Appendix \ref{sec:app-models}.

\subsection{Prompting LLMs}

To explore the potential of recent instruction-tuned LLMs for historical text classification, we also experimented with Llama 3.1 (8B) and Qwen 2 (7B). These are general-purpose models trained on very large collections of texts in multiple languages and domains, which gives them broad linguistic and contextual knowledge. In addition, they are instruction-tuned, which means that they have been further adapted to follow natural language instructions (e.g. "classify this text..."). In our study, we used a prompt-based classification strategy in a zero-shot setting: the models were asked to decide whether a text contained invective without being trained on task-specific examples beforehand. The prompt formulation was derived from the working definition of invective language presented in Section~\ref{historical-invectives} and then operationalized in a concise instruction format. It emphasizes the essential aspects of invectivity while ensuring clarity and brevity for the models. Specifically, for both models the following instruction prompt was used:

\begin{tcolorbox}[colback=gray!10, colframe=gray!50, left=2mm, right=2mm, top=1mm, bottom=1mm, boxrule=0.5pt]
\textit{Decide whether the given sentence contains invectives or not. 
Invectives include all utterances that have the potential to disparage an opponent person or group in the context of the given sentence. 
They may target an opponent directly or indirectly, and can be literal or metaphorical. 
Answer with only one word: Invective or Non-invective.}
\end{tcolorbox}

\section{Results}

\subsection{Classification results}

Table~\ref{tab:model-performance} shows that the fine-tuned BERT-based models clearly outperform the LLMs (Llama 3 and Qwen 2). The best results come from MacBERTh, which reaches 0.89 F1. This makes sense as MacBERTh was pre-trained on historical language, so it is better suited to the kind of texts in our dataset. The other BERT-based models do reasonably well too, but not as strongly. BERT-base (0.83 F1) performs slightly worse than MacBERTh, likely because it was pretrained on modern English and therefore does not capture historical wording and style as well. XLM-R (0.82 F1) shows similar performance. 
By contrast, the LLMs score lower overall (0.76 and 0.70 F1). The gap between the two is also notable. Llama 3 performs considerably better than Qwen 2, which may reflect differences in pretraining data and alignment strategies. The superior performance of the BERT-based models likely reflects the full fine-tuning on the task which allows to internalize the lexical and contextual cues that distinguish invective from non-invective language. In contrast, the LLMs were used in a prompted zero-shot setting and rely on general-purpose pretraining. 

\begin{table}[!ht]
\centering
\resizebox{0.7\columnwidth}{!}{%
\begin{tabular}{llccccccccc}
\toprule
\textbf{Model} & \textbf{Prec.} & \textbf{Rec.} & \textbf{F1} \\
\midrule
MacBERTh & \textbf{0.89} & \textbf{0.88} & \textbf{0.89} \\
BERT-base & 0.84 & 0.82 & 0.83 \\
XLM-R & 0.83 & 0.81 & 0.82 \\ \midrule
Llama-3.1-8B & 0.79 & 0.75 & 0.76 \\
Qwen2-7B & 0.71 & 0.76 & 0.70 \\
\bottomrule
\end{tabular}
}
\caption{Classification performance of our tested models, macro averages.}
\label{tab:model-performance}
\end{table}

\subsection{Error Analysis}
\label{sec:error-analysis}

To better understand the strengths and weaknesses of the different models, we carried out both quantitative and qualitative error analyses. This sheds light on systematic biases in prediction and those kinds of cases which remain challenging across models.  

\subsubsection{Quantitative analysis}

\begin{figure}[h!]
    \centering
    \includegraphics[width=0.8\columnwidth]{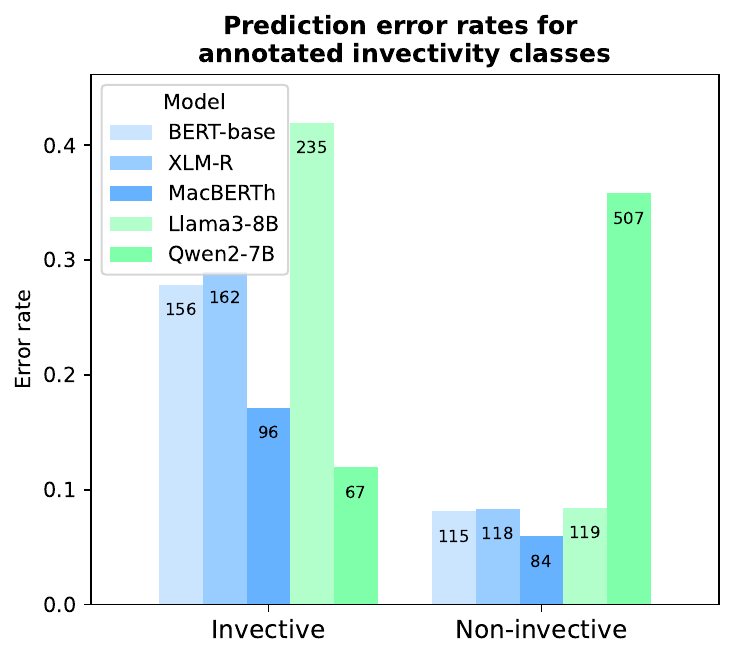}
    \caption{Prediction error rates for the tested models (with BERT-based in blue and the LLMs in green) for the different annotated invectivity classes. Each bar represents the proportion of misclassifications within a given class.}
    \label{fig-errors-invective}
\end{figure}

Prediction error rates reveal notable differences in model behavior for invective and non-invective text, as visualized in Figure \ref{fig-errors-invective}. MacBERTh exhibits the most balanced performance, with low error rates for both invective (17.1\%) and non-invective text (5.9\%). BERT-base and XLM-R show consistent and moderately well error rates across both classes. Qwen2 achieves the lowest error rate for invective content (12.0\%) but performs poorly on non-invective text (35.8\%).  Conversely, Llama3 struggles with invective (42.0\%) but performs moderately on non-invective text (8.4\%). The LLM results thus indicate a strong bias toward one class: Qwen2 tends to overpredict invective, while Llama3 tends to underpredict it.

\begin{figure}[h!]
    \centering
    \includegraphics[width=0.8\columnwidth]{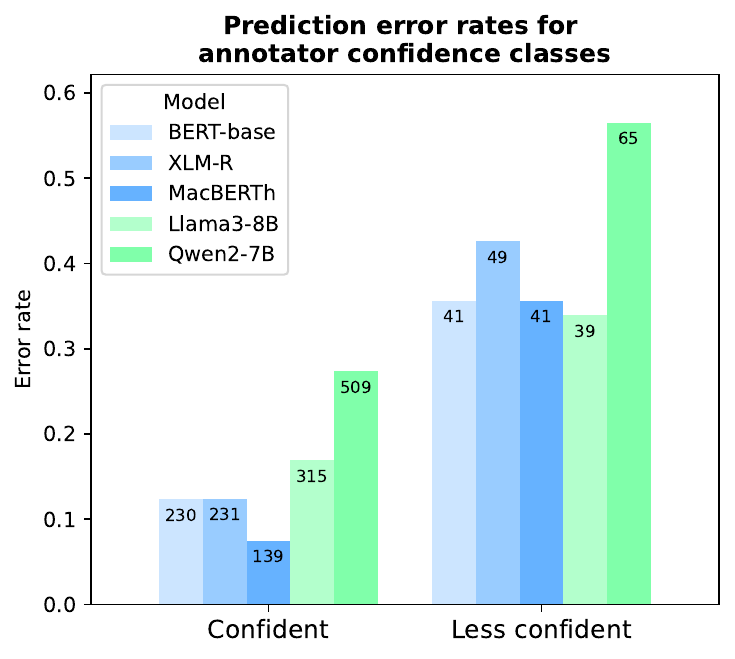}
    \caption{Prediction error rates for the tested models (with BERT-based in blue and the LLMs in green) for the different annotator confidence classes. Each bar represents the proportion of misclassifications within a given class.}
    \label{fig-errors-confidence}
\end{figure}

Figure \ref{fig-errors-confidence} displays the prediction error rates for different models on text annotated with confident and less confident labels. Error rates increased substantially for less confident annotations. Llama3 (33.9\%), BERT-base and MacBERTh (both 35.7\%) show moderate performance and XLM-R (42.6\%) and Qwen2 (56.5\%) struggle the most. For these examples, the model errors mirror human uncertainty which suggests that the challenge arises from the inherent difficulty of such cases rather than from the models. 

\subsubsection{Qualitative analysis}

We conducted an initial examination of the sentences that were incorrectly classified by the models and subsequently present first observations regarding potential sources of errors. However, a more comprehensive analysis is required to validate these impressions. 

One of the challenges to classify non-invective sentences correctly seems to be that the data may include negatively connotated language, which depending on the context does not necessarily convey an invective meaning. There is often only a fine line between the mere reference to sinful conduct or heretical belief and its denunciation. For example, this is the case in many devotional passages:

\begin{quote} 
    “\textit{i poore miserable and wretched creatur, knowledge, and confesse before thy fatherly mercy, that i, lyke the ryottous sonne, haue despised and not regarded all the same thy fatherlye loue and trust}”\footnote{Heyden, C. v. d. (1550). \href{https://www.proquest.com/books/bryefe-summe-whole-byble-chrystian-instruction-al/docview/2240927568/se-2}{A bryefe summe of the whole byble.}  London. 
    }
\end{quote}  

Vice versa, we also find positively connotated and \textit{"not" + negative term} constructions in sentences which have been misclassified as \textit{non-invective} by the models. 



Metaphors pose a challenge in both directions: either they go undetected, causing a sentence to be misclassified as non-invective, or literal statements are mistakenly interpreted as invective metaphors and classified as such. They often require a deeper understanding of the context of the historical sources. For example, the image of the wolf at that time is a common derogatory term referring to Catholics: 

\begin{quote} 
    “\textit{let not eche rauenynge woulfe that cometh wyth a shepehoke in hys hande be receued as a shepherde.}”\footnote{Anonymous. (1546). \href{https://www.proquest.com/books/supplication-poore-commons-whereunto-is-added/docview/2248571291/se-2}{A supplication of the poore commons Whereunto is added the supplication of beggers.} London.}
\end{quote}  

Our preliminary insights reinforce the importance of qualitatively examining errors. 

\section{Discussion \& Conclusion}

Our study presents the InviTE corpus, a sentence-level corpus of invective language in Early Modern English. Beyond introducing the corpus itself, we outlined a workflow that combines historical and computational perspectives, from data selection and pre-processing to annotation design and application. At each step, our methods are closely aligned with the implications of the EModE sources at the base of the corpus and their historical context. This interdisciplinary approach enabled us to operationalize theories of invective language for computational purposes. In doing so, we address challenges such as subjectivity in annotation and the sparsity of invective language in the data. By integrating confidence scores and refining categories through iterative annotation, we established a methodology that can inform similar efforts in historical and domain-specific corpus creation.

Turning to our computational experiments, we showed that fine-tuned encoder models outperform much larger LLMs in zero-shot classification. This finding highlights the continuing importance of supervised adaptation when working in specialized domains, even in the era of instruction-tuned LLMs. In addition, pre-training on the right kind of language matters. MacBERTh’s superior performance demonstrates that pre-training on historical corpora offers a strong advantage for classification tasks involving non-contemporary language. Nevertheless, it is noteworthy that both LLMs achieved competitive performance without any task-specific training. This underlines the adaptability of modern LLMs, which is especially valuable when fine-tuning is not feasible, for instance due to the resource demands of manually labeling large amounts of (historical) data. At the same time, the results suggest that further gains are likely achievable through targeted post-training or fine-tuning to better align the models with the specific domain and task. 

In presenting the InviTE corpus, we \textit{invite} further work at the intersection of History and NLP. 



\section*{Limitations}

Despite the promising results presented, this study is subject to several limitations that should be considered. 
The method used to compile the InviTE corpus, particularly the targeted addition of invective sentences based on keyword searches, introduces a selection bias. While this step was necessary to mitigate the inherent sparsity of invective language and to create a more balanced dataset suitable for computational model training, it means that the corpus may not reliably be used for quantitative historical analyses of e.g. invective prevalence across time periods or author groups. The proportional representation of invective versus non-invective language does not reflect the original distribution in EModE texts.
Besides, the IAA metrics we rely on do not account for interdependencies of the annotation categories we applied. As \citet{reiter_erstellung_2020} notes, research endeavors within the digital humanities often require complex annotation schemes and no robust solutions have been developed to represent these dependencies within IAA values so far.

Regarding the evaluation of model performances, the only evaluated LLMs in a zero-shot setting. Recent research has shown that few-shot learning approaches can substantially improve performance for instruction-tuned models. Exploring these strategies could enhance invective detection and is left for future work.
Additionally, our computational experiments focused on the main binary invective classification, the InviTE corpus also contains additional annotation categories capturing more fine-grained aspects of invective language. These richer annotations were not leveraged in the current study, but they present promising directions for future research.

\section*{Ethical Statement}
Studying (historical) invectives inevitably comes with reproducing biases. While the sources data back to the 16\textsuperscript{th} century, inter- and intrareligious invectives remain a pressing issue to this day. Therefore, it is crucial to ensure that the data is only used to for scientific purposes. 

Moreover, although LLMs prove to be powerful tools for further interdisciplinary research, the cost-benefit ratio remains to be assessed individually, with due consideration given to resource availability as well as questions concerning the hosting of the models and data protection requirements.

\section*{Acknowledgments}

The authors acknowledge financial support by the project ``SAIL: SustAInable Life-cycle of Intelligent Socio-Technical Systems'' (Grant ID NW21-059A), which is funded by the program ``Netzwerke 2021'' of the Ministry of Culture and Science of the State of Northrhine Westphalia, Germany.

\bibliography{custom}

\appendix

\section{Keywords}
\label{sec:app-keywords}
Keywords used for data selection (an asteriks denotes a wildcard in the search process): agaynst, babling*/ babbling*, backbit*, blind*, cruel*, euil*, deceiu*, destroy*, destruct*, deuil*/ devil*, false, fault*, fornicat*, fool*, foul*, harlot*, heretic*, hypocrite*, idolat*, infidel*, invect*/inuect*, liar*, papist*, persecut*, persuad*/ perswad*, persuas*, pop*, sin*/ syn*, unbeleev*/unbeleeu*, wicked*. 

\section{Annotation example}
\label{sec:app-ann}

An example of an annotated sentence is presented in Table \ref{tab:ann-example}.

\begin{table*}[!ht]
    \centering
    \resizebox{0.8\textwidth}{!}{%
    \begin{tabular}{p{6cm}cccc}
        \textbf{Sentence} & \textbf{Invectivity} & \textbf{Type} & \textbf{Target} & \textbf{Confidence}\\ \midrule
         in submittyng your selues to that fylthye beast of rome, and in receauyng the stinckyng i dolatrous masse, by the whiche you haue destreyed an inaumerable sorte of people? & invective & metaphorical & religious confession & confident\\ \bottomrule
    \end{tabular}
    }
    \caption{Example sentence annotated according to the pre-defined categories.}
    \label{tab:ann-example}
\end{table*}

\section{Model implementation details}
\label{sec:app-models}

For the BERT-based models, training was performed for 5 epochs with a batch size of 32 and a learning rate of $2 \times 10^{-5}$, using the AdamW optimizer and a linear learning rate scheduler. The best model checkpoint was selected based on the lowest average training loss on the validation set. 

We used \texttt{StratifiedKFold} from \texttt{sklearn.model\_selection} to generate folds and batches that preserve the class distribution within each fold or batch. 

All pretrained models and tokenizers were loaded from the Transformers library:

\begin{itemize}
    \item BERT-base: \url{https://huggingface.co/google-bert/bert-base-uncased}
    \item XLM-Roberta-Large:  \url{https://huggingface.co/FacebookAI/xlm-roberta-large}
    \item MacBERTh: \url{https://huggingface.co/emanjavacas/MacBERTh}
    \item Llama 3.1 (8B) Instruct - \url{https://huggingface.co/meta-llama/Llama-3.1-8B-Instruct} 
    \item Qwen2 (7B) Instruct: \url{https://huggingface.co/Qwen/Qwen2-7B-Instruct}
\end{itemize}

All the code used for the computational experiments is available via \url{https://github.com/SanneHoeken/InviTE-experiments}.


\end{document}